\pdfoutput=1

\documentclass[11pt]{article}
\usepackage{tabularx}
\usepackage{booktabs}
\usepackage{multirow}
\usepackage{multicol}
\usepackage[preprint]{acl}

\usepackage{times}
\usepackage{latexsym}

\usepackage[T1]{fontenc}

\usepackage[utf8]{inputenc}

\usepackage{microtype}

\usepackage{inconsolata}

\usepackage{graphicx}

\usepackage{enumitem}
\usepackage{amsmath}
\usepackage{calc}

\usepackage{numprint}

\npdecimalsign{.}
\npthousandsep{,}

\usepackage{makecell}

\usepackage{siunitx}
\usepackage{listings}
\usepackage{stfloats}

\title{Beyond Repetition: Text Simplification and Curriculum Learning for Data-Constrained Pretraining}

\author{
    Matthew Theodore Roque$^{\ast}$ \and Dan John Velasco$^{\ast}$ \\
    Samsung R\&D Institute Philippines\\
  \texttt{\{roque.mt,dj.velasco\}@samsung.com} \\
  \small{\textbf{$^{\ast}$Equal Contribution}} \\
}

\begin{document}
\maketitle
\begin{abstract}
Most studies on language model pretraining focus on large datasets, leaving open questions about optimization in data-constrained settings. In such settings, the effects of training data order and of including alternative versions of the same text remain underexplored. We address this by studying curriculum learning in pretraining, focusing on text-complexity ordering and data augmentation via simplification. We ask: (1) Does simplifying texts enhance representation quality more than reusing the original data? and (2) Does ordering data by text complexity yield better representations? To answer, we build on a pair of parallel corpora where human-written paragraphs are aligned with LLM-simplified variants, and test four data schedules: repeated exposure, low-to-high complexity, high-to-low, and interleaved. We analyze models' representation quality from a sample efficiency perspective via fine-tuning, as well as its zero-shot performance on linguistic knowledge, entity tracking, world knowledge, and commonsense reasoning. Our findings show that adding simplified data improves fine-tuning and zero-shot performance over a repeated-exposure baseline: smaller models benefit from low-to-high complexity, while larger models perform better with interleaved ordering.
\end{abstract}

\begin{figure}[t]
    \centering
    \includegraphics[width=\columnwidth]{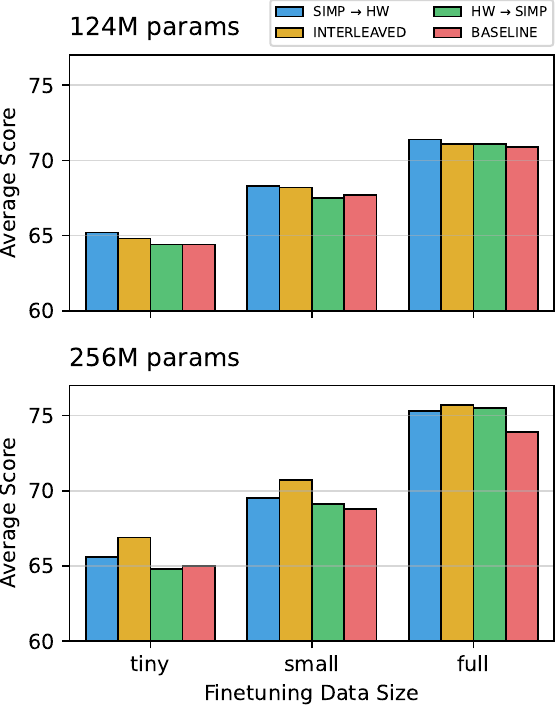}
    \caption{Average score on six language tasks by curriculum and fine-tuning data size. \textbf{(Top)} The 124M model benefits from \texttt{SIMP} $\rightarrow$ \texttt{HW} curriculum, suggesting smaller models gain from warming up on simpler text. \textbf{(Bottom)} The 256M model benefits from \texttt{INTERLEAVED}, favoring balanced exposure. All but \texttt{HW} $\rightarrow$ \texttt{SIMP} outperform \texttt{BASELINE} in sample efficiency.}
    \label{fig:main-results}
\end{figure}

\section{Introduction}
Scaling studies show that language model performance improves predictably with more data, parameters, and compute \citep{kaplan2020scaling, hoffmann2022training}. However, these studies typically assume that the amount of unique pretraining data is effectively unlimited \citep{muennighoff2025scalingdataconstrainedlanguagemodels}. In practice, pretraining often faces data-constrained settings where continued exposure to the same corpus is unavoidable. Under such conditions, two factors remain underexplored in modern decoder-only pretraining: (1) the order in which training data is presented, and (2) the use of simpler versions of the same text.

We examine these factors concretely. For training data order, we focus on \textbf{coarse-grained text complexity ordering}—presenting a simple corpus before a complex one—while keeping the core content constant, following the intuition of curriculum learning \citep{bengio2009curriculum}. We define "simple" texts as those that use higher-frequency vocabulary and have shallower syntactic structures. 

\pagebreak
Our experiments use a high-quality English dataset where human-written paragraphs are paired with simplified counterparts produced by an LLM. This corpus was introduced and validated in concurrent work \citep{velasco2025rethinking}, which demonstrates that simplification reduces surface-level complexity (sentence length, syntactic depth, lexical diversity) while preserving semantic content. Here, we do not revisit corpus construction in detail; instead, we leverage it to test how simplified data and ordering strategies interact under a fixed training budget.

\paragraph{We evaluate four data schedules:}
\begin{itemize}
  \item \texttt{BASELINE}: repeated exposure to human-written text.
  \item \texttt{INTERLEAVED}: human-written and simplified paragraphs uniformly mixed.
  \item \texttt{SIMP$\rightarrow$HW}: training first on simplified, then human-written text (curriculum).
  \item \texttt{HW$\rightarrow$SIMP}: training first on human-written, then simplified text (anti-curriculum).
\end{itemize}
All other training variables (architecture, tokenizer, context length, optimizer) are held constant across schedules, isolating the effect of text complexity and order.

\paragraph{Research questions.}
We ask two core questions:
\begin{enumerate}
  \item[(1)] In data-constrained settings, does replacing repeated exposure with simplified text improve representation quality?  
  \item[(2)] Does ordering data by text complexity—simple to complex versus interleaved—yield better downstream and zero-shot performance?
\end{enumerate}

\paragraph{Contributions.}
Our contributions are threefold:
\begin{enumerate}
  \item[(1)] We provide the first controlled study of how text simplification and curriculum scheduling interact in data-constrained pretraining.  
  \item[(2)] We evaluate these schedules across fine-tuning and zero-shot tasks, covering linguistic knowledge, entity tracking, world knowledge, and commonsense reasoning.  
  \item[(3)] We show that simplified data generally improves performance over repeated exposure, that smaller models benefit from simple-to-complex curricula, and that larger models favor balanced exposure via interleaving.
\end{enumerate}

\section{Related Work}
\label{sec:related}

\paragraph{Data-constrained Pretraining and Synthetic Data.}
Most scaling studies assume unlimited data. In data-constrained settings, \citet{muennighoff2025scalingdataconstrainedlanguagemodels} shows that training up to four epochs of repeated data is just as good as unique data, but further repetition offers no benefit. Recent works address the "data wall" either by using LLMs to \textit{generate} synthetic data \cite{gunasekar2023textbooksneed,benallal2024cosmopedia} or to \textit{rewrite} existing data, broadly across general domains \citep{maini-etal-2024-rephrasing,su-etal-2025-nemotron,nguyen2025recyclingwebmethodenhance,datologyai2025beyondweblessonsscalingsynthetic} and in specific domains such as math, code, and clinical text \citep{fujii2025rewritingpretrainingdataboosts,liu-nguyen-2024-rephrasing}. Rewriting into simpler forms remains underexplored; we investigate this setting.

\paragraph{Curriculum Learning.}
Humans learn better when examples follow a meaningful order (e.g., simple to complex). \citet{bengio2009curriculum} formalized this as Curriculum Learning (CL), training neural networks on data of gradually increasing complexity. The opposite, anti-curriculum, has sometimes matched or outperformed CL \citep{kocmi-bojar-2017-curriculum,zhang2018empiricalexplorationcurriculumlearning}. Difficulty metrics vary by application: in code language models, \citet{nair-etal-2024-curriculum} used software engineering metrics, while in machine translation, \citet{zhou-etal-2020-uncertainty} measured sequence uncertainty via language model entropy. Recent studies applied CL to pretraining language models: \citet{tsvetkov-etal-2016-learning} examined linguistically inspired measures such as age of acquisition, while \citet{oba-etal-2023-babylm} measured complexity via dependency tree depth. A recent large-scale study by \citet{zhang2025randomsamplingefficientlanguage} finds that using text complexity metrics such as Flesch Reading Ease, Lexical Diversity, and Compression Ratio can accelerate convergence and modestly outperform random shuffling.

However, most CL studies in pretraining conflate language and content complexity, failing to isolate the effect of ordering by language complexity. Most pretraining studies also assume large data volumes, leaving open the question of how to advance pretraining in data-constrained settings beyond repeated exposure. This work uniquely addresses this gap, testing whether LLM-based simplification and coarse-grained text complexity ordering improve representational quality beyond repeated exposure to the original data.

\section{Methodology}
\label{sec:method}

\subsection{Data}
We reuse the two parallel corpora from \citet{velasco2025rethinking}, derived from a 2B-token subset of FineWeb-Edu (ODC-By 1.0; \citealp{penedo2024fineweb}). The human-written corpus (\texttt{HW}) and its simplified variant (\texttt{SIMP}) are aligned at the paragraph level: each paragraph in \texttt{HW} has a corresponding simplification in \texttt{SIMP} produced by an LLM, with paragraphs that fail basic length/formatting checks symmetrically removed from both sides to preserve one-to-one alignment. After filtering, token counts are approximately 2.00B (\texttt{HW}) and 1.71B (\texttt{SIMP}). 

In brief, \texttt{SIMP} reduces surface-level complexity (shorter sentences, shallower syntax, more frequent vocabulary) while preserving core content; \citet{velasco2025rethinking} report readability and lexical/syntactic metrics alongside semantic-similarity checks that validate this property. Our study focuses on how to use these corpora under a fixed budget: repeated exposure vs. augmentation with simplifications and interleaving vs. ordered curricula. All other variables (architecture, tokenizer, context length, optimizer) are held constant across schedules.

\subsection{Schedules}
We compare four data schedules that differ only in the order of data presented to the model. Unlike most curriculum strategies that manipulate complexity per example (fine-grained), our schedules are coarse-grained, adjusting complexity at the corpus level rather than per example. The four schedules are as follows:
\begin{itemize}
  \item \texttt{BASELINE}: two epochs of \texttt{HW} (simulating data-constrained scenarios).
  \item \texttt{INTERLEAVED}: \texttt{HW} and \texttt{SIMP} are uniformly interleaved, preserving each corpus’s within-source order (simulating random shuffling in a more balanced way).
  \item \texttt{SIMP$\rightarrow$HW}: concatenation of \texttt{SIMP} and \texttt{HW} (simulating standard curriculum).
  \item \texttt{HW$\rightarrow$SIMP}: concatenation of \texttt{HW} and \texttt{SIMP} (simulating anti-curriculum).
\end{itemize}

Each training example is a single paragraph, and both \texttt{HW} and \texttt{SIMP} corpora are perfectly parallel, containing the same number of paragraphs. Within each source, paragraph order is fixed across all schedules, ensuring differences arise only from the sequence in which the two sources are presented. This design isolates the effect of data ordering by text complexity and the presence of LLM-rewritten text, while controlling for content coverage and total training steps.

\subsection{Model and Training}
We use 124M and 256M parameter causal language models based on the design principles of MobileLLM \citep{liu2024mobilellmoptimizingsubbillionparameter}, adopting a deep-and-thin architecture with SwiGLU activations \citep{shazeer2020gluvariantsimprovetransformer}, grouped-query attention \citep{ainslie-etal-2023-gqa}, and without embedding weight sharing for better comparability with contemporary decoder-only models. Each model has 30 transformer layers with 9 attention heads (3 key–value heads per layer) and embedding dimensions of 576 (124M) and 846 (256M). We refer to the configurations as 124M/256M by convention; the total parameter counts including embeddings are approximately 143M and 283M, respectively.

All corpora are tokenized using the LLaMA-2 BPE tokenizer \citep{touvron2023llama2openfoundation} with a 32,000-token vocabulary. Training examples are individual paragraphs, with no concatenation or sequence packing to control for total training steps. Inputs are right–padded to 512 tokens with the EOS token.

Optimization uses AdamW \citep{loshchilov2019decoupledweightdecayregularization} with default hyperparameters, a peak learning rate of $3\mathrm{e}{-4}$ linearly decayed over training, 5\% warm-up, and no dropout. The effective batch size is 256 (8 examples per GPU × 8 GPUs × 4 gradient accumulation steps). Training is conducted in FP16 mixed precision on 8× NVIDIA P100 GPUs, without gradient checkpointing. All experiments use PyTorch with Hugging Face Transformers and Distributed Data Parallel (DDP), with a fixed random seed of 42 for data shuffling and parameter initialization.


\subsection{Evaluation Setup}

We evaluate each schedule under two complementary regimes to capture both transferable language understanding after fine-tuning and generalization without further task-specific training.

\paragraph{Fine-tuning on NLU tasks.}
We fine-tune the pretrained models on a subset of the BabyLM evaluation pipeline’s natural language understanding (NLU) tasks, using the preprocessed training and validation splits provided therein \citep{charpentier2025babylmturns3papers}. 
From the original suite, we exclude:
\begin{itemize}
    \item \textbf{MultiRC}, due to its substantially larger size and because preliminary experiments showed none of our pretraining setups outperformed training from scratch (58–59 points).
    \item \textbf{WSC}, due to its much smaller dataset size and the high variance (4–12 points) observed across seeds under different fine-tuning budgets.
\end{itemize}
The resulting set of tasks is: BoolQ, MNLI, MRPC, QQP, and RTE. Each task is framed as classification, with paired-input tasks (e.g., premise-hypothesis) concatenating the two sequences with a separator token. 
Fine-tuning is performed with all model parameters trainable. 
Batch sizes are set by memory constraints: 2 examples per GPU (effective batch size 16) for BoolQ, and 8 examples per GPU (effective batch size 64) for all other tasks. 
For each task, we search over learning rates $\{1\mathrm{e}{-4}, 5\mathrm{e}{-5}, 2\mathrm{e}{-5}, 1\mathrm{e}{-5}, 5\mathrm{e}{-6}\}$ and epochs $\{1, 2, 3, 4, 5\}$. 
Model selection is based on the highest validation score according to the task’s standard metric (accuracy for BoolQ, MNLI, and RTE; F1 for MRPC and QQP). 
All results are averaged over three runs with different random seeds.

In addition to this \textbf{Full} fine-tuning setup, we introduce two smaller-scale, class-balanced variants to test whether having more downstream task examples diminishes the influence of pretraining differences. 
For each task, we identify the class with the fewest examples in the original training split, then construct:
\begin{enumerate}
    \item \textbf{Small} dataset, in which all classes contain exactly half of the least-represented class size, and
    \item \textbf{Tiny} dataset, in which all classes contain exactly one quarter of the least-represented class size.
\end{enumerate}
This ensures that all classes are equally represented and that dataset size is systematically reduced across tasks.

These smaller datasets are fixed across runs so that each of the three seeds per setup uses the same subset of examples, with only model initialization differing. 
The same hyperparameter sweeps are applied to these reduced datasets as in the full-data setup.

\begin{table}[t]
\centering
\footnotesize
\begin{tabularx}{\columnwidth}{@{}Xcrrr@{}}
\toprule
\textbf{Task} & 
\multicolumn{1}{r}{\textbf{Classes}} & 
\multicolumn{1}{r}{\textbf{Full}} & 
\multicolumn{1}{r}{\textbf{Small}} & 
\multicolumn{1}{r}{\textbf{Tiny}} \\
\midrule
MNLI   & 3 & 10{,}000 & 4{,}911 (49\%) & 2{,}454 (25\%) \\
MRPC   & 2 & 3{,}668  & 1{,}194 (33\%) & 596 (16\%) \\
BoolQ  & 2 & 9{,}427  & 3{,}552 (38\%) & 1{,}776 (19\%) \\
RTE    & 2 & 2{,}490  & 1{,}240 (50\%) & 620 (25\%) \\
QQP    & 2 & 10{,}000 & 3{,}662 (37\%) & 1{,}830 (18\%) \\
\bottomrule
\end{tabularx}
\caption{
Number of training examples per NLU task in the \textbf{Full}, \textbf{Small}, and \textbf{Tiny} fine-tuning setups. 
Small and tiny datasets are class-balanced subsets, with percentages shown relative to the full dataset. 
Percentages vary across tasks because subset sizes are determined relative to the least-represented class in each dataset.
}
\label{tab:nlu-data-sizes}
\end{table}

\paragraph{Zero-shot evaluation.}
Using the LM Evaluation Harness \citep{eval-harness}, we assess models trained under different curricula via zero-shot performance on a suite of multiple-choice benchmarks, grouped into three capability areas:
\begin{itemize}
    \item \textbf{Linguistic knowledge:} BLiMP \citep{warstadt2020blimp} and BLiMP Supplement \citep{charpentier2025babylmturns3papers} for syntactic and morphological phenomena.
    \item \textbf{Discourse and world knowledge:} Entity Tracking \citep{kim-schuster-2023-entity} for reference consistency, and MMLU \citep{hendrycks2021measuringmassivemultitasklanguage} and EWoK \citep{ivanova2024elements} for factual knowledge.
    \item \textbf{Commonsense reasoning:} ARC \citep{allenai:arc}, HellaSwag \citep{zellers2019hellaswag}, PIQA \citep{Bisk2020}, Social IQa \cite{sap-etal-2019-social}, and OpenBookQA \cite{mihaylov-etal-2018-suit} for reasoning about everyday scenarios.
\end{itemize}

All zero-shot tasks are formatted according to their official specifications. For each candidate answer, we compute the sum of log-probabilities of its tokens given the prompt and select the option with the highest score. Accuracy is reported for all tasks. Evaluation is deterministic and performed on a single NVIDIA P100 GPU.

\section{Results and Discussion}
\label{sec:results}
Across fine-tuning and zero-shot setups, outcomes hinge on two factors: augmenting with simplified data versus repeated exposure, and ordering of data from simplified to complex versus interleaving.



\begin{table}[t]
\centering
\footnotesize
\sisetup{round-precision=1,round-mode=places,detect-weight,detect-family,table-number-alignment=center}
\begin{tabularx}{\columnwidth}{@{}Xl *{3}{S[table-format=2.1]}@{}}
\toprule
\textbf{Model} & \textbf{Full} & \textbf{Small} & \textbf{Tiny} \\
\midrule
\multicolumn{4}{@{}l}{\textbf{124M}} \\
\texttt{BASELINE}              & 70.9 & 67.7 & 64.4 \\
\texttt{INTERLEAVED}           & 71.1 & 68.2 & 64.8 \\
\texttt{SIMP} $\rightarrow$ \texttt{HW} & \bfseries 71.4 & \bfseries 68.3 & \bfseries 65.2 \\
\texttt{HW} $\rightarrow$ \texttt{SIMP} & 71.1 & 67.5 & 64.4 \\
\midrule
\multicolumn{4}{@{}l}{\textbf{256M}} \\
\texttt{BASELINE}              & 73.9 & 68.8 & 65.0 \\
\texttt{INTERLEAVED}           & \bfseries 75.7 & \bfseries 70.7 & \bfseries 66.9 \\
\texttt{SIMP} $\rightarrow$ \texttt{HW} & 75.3 & 69.5 & 65.6 \\
\texttt{HW} $\rightarrow$ \texttt{SIMP} & 75.5 & 69.1 & 64.8 \\
\bottomrule
\end{tabularx}
\caption{Macro-average across five NLU tasks (BoolQ, MNLI, MRPC-F1, QQP-F1, RTE) under three fine-tuning budgets: \textbf{Full}, \textbf{Small}, and \textbf{Tiny}. Best per model size per budget in bold. At 124M, \texttt{SIMP}$\rightarrow$\texttt{HW} leads across all budgets (+0.5–0.8 vs. \texttt{BASELINE}); at 256M, \texttt{INTERLEAVED} leads across all budgets (+1.8–1.9), indicating size- and budget-dependent preferences. Per-task breakdowns with mean ± std. dev. are in Appendix \ref{sec:per-task-tables}.}
\label{tab:nlu-summary}
\end{table}

\subsection{Simplification vs. Repetition}
We ask whether replacing a second pass over the human-written corpus with LLM-based simplifications improves pretraining. The primary comparison is \texttt{BASELINE} (two epochs of \texttt{HW}) vs.\ \texttt{INTERLEAVED}, which uniformly mixes \texttt{HW} and \texttt{SIMP}.

\paragraph{Fine-tuning evaluation.}

At 124M, adding \texttt{SIMP} yields small, consistent gains over \texttt{BASELINE} when mixed (\texttt{INTERLEAVED}: +0.2, +0.5, +0.4 on \textbf{Full}/\textbf{Small}/\textbf{Tiny}). At 256M, the benefits of including \texttt{SIMP} are larger overall in the mixed setting (\texttt{INTERLEAVED}: +1.8/+1.9/+1.9). These advantages grow as fine-tuning data decreases: \texttt{INTERLEAVED} degrades less than \texttt{BASELINE} on Small and Tiny setups, suggesting simplification is most beneficial when data is scarce, improving sample efficiency.

\paragraph{Zero-shot evaluation.}
Across linguistic, discourse, and commonsense benchmarks, introducing \texttt{SIMP} tends to be neutral to positive at 124M and more evidently positive at 256M, with the largest single gain appearing on entity tracking when mixing (\texttt{INTERLEAVED} vs.\ \texttt{BASELINE}: $\approx+4.9$ points at 256M, Table~\ref{tab:ling-benchmarks}). Other tasks show smaller, sometimes mixed, changes, so the aggregate trend favors simplification but with task-family variability (Table~\ref{tab:reasoning-benchmarks}).

\paragraph{Summary.}

Compared to repeating \texttt{HW}, in-domain augmentation via LLM simplification yields modest gains at 124M and larger gains at 256M. While we do not claim causality, a plausible explanation is that the additional capacity of the 256M model may better absorb paraphrastic variety, making mixed (\texttt{INTERLEAVED}) exposure yield small but consistent improvements in both fine-tuned and zero-shot settings. We consider this hypothesis specific to our training setup rather than a general rule.

While repetition underperforms simplification, it remains attractive given its zero generation cost and competitive results, consistent with findings by \citet{muennighoff2025scalingdataconstrainedlanguagemodels}. \textbf{Our results complement rather than replace repeated exposure: since performance saturates beyond four epochs, text simplification offers a way to extend these gains further in data-constrained settings.}

\begin{table*}[t]
\centering
\footnotesize
\begin{tabularx}{\textwidth}{@{}Xccccccc@{}}
\toprule
\textbf{Model} & 
\multicolumn{1}{r}{\textbf{arc\_chl}} & 
\multicolumn{1}{r}{\textbf{arc\_e}} & 
\multicolumn{1}{r}{\textbf{hellaswag}} & 
\multicolumn{1}{r}{\textbf{openbookqa}} & 
\multicolumn{1}{r}{\textbf{piqa}} & 
\multicolumn{1}{r}{\textbf{social\_iqa}} & 
\multicolumn{1}{r}{\textbf{Avg.}} \\
\midrule

\textbf{124M} & & & & & & & \\
\texttt{BASELINE}            & 22.0 & \textbf{35.9} & \textbf{28.4}  & 14.0 & 57.1 & 36.0 & 34.28 \\
\texttt{INTERLEAVED}            & \textbf{23.4} & 34.6 & \textbf{28.4}  & 15.4 & \textbf{58.4} & 35.7 & \textbf{34.50}\\
\texttt{SIMP} $\rightarrow$ \texttt{HW}  & 23.1 & 33.7 & 28.0 & \textbf{17.0} & 57.3 & 36.1 & 34.42\\
\texttt{HW} $\rightarrow$ \texttt{SIMP}  & 20.4 & 33.5 & 28.1  & 14.8 & 57.6 & \textbf{36.9} & 34.18\\
\midrule
\textbf{256M} & & & & & & & \\
\texttt{BASELINE}            & 23.1 & \textbf{37.8} & 27.7& 16.0 & 57.3 & 35.2 & 34.80\\
\texttt{INTERLEAVED}            & 22.7 & 36.9 & 28.6 & 16.2 & 56.8 & \textbf{36.6} & 35.02\\
\texttt{SIMP} $\rightarrow$ \texttt{HW}  & 21.8 & 35.5 & \textbf{29.1}  & \textbf{18.2} & \textbf{58.4} & 36.5 & \textbf{35.54}\\
\texttt{HW} $\rightarrow$ \texttt{SIMP}  & \textbf{24.7} & 34.7 & 28.2 & 16.6 & 56.4 & 35.5 & 34.28\\
\bottomrule
\end{tabularx}
\caption{Zero-shot accuracy on commonsense reasoning benchmarks (ARC-Challenge, ARC-Easy, HellaSwag, OpenBookQA, PIQA, Social IQa). “Avg.” is the mean across tasks. At 124M, \texttt{INTERLEAVED} yields the highest average by a small margin; at 256M, \texttt{SIMP} $\rightarrow$ \texttt{HW} attains the best average, driven by gains on HellaSwag, OpenBookQA, and PIQA, while \texttt{INTERLEAVED} leads on Social IQa and \texttt{HW} $\rightarrow$ \texttt{SIMP} peaks on ARC-Challenge.}
\label{tab:reasoning-benchmarks}
\end{table*}

\begin{table}[t]
\centering
\footnotesize
\begin{tabularx}{\columnwidth}{@{}Xccccc@{}}
\toprule
\textbf{Model} & 
\multicolumn{1}{r}{\textbf{blimp}} & 
\multicolumn{1}{r}{\textbf{supp}} & 
\multicolumn{1}{r}{\textbf{ewok}} & 
\multicolumn{1}{r}{\textbf{entity}} &  
\multicolumn{1}{r}{\textbf{mmlu}} \\
\midrule

\textbf{124M} & & & & &\\
\texttt{BASELINE}            & 71.8 & 61.9 & 53.9 & 22.4 & 24.7\\
\texttt{INTERLEAVED}            & 72.3 & 63.6 & \textbf{55.5} & 28.1 & \textbf{24.9}\\
\texttt{SIMP} $\rightarrow$ \texttt{HW}  & \textbf{72.4} & \textbf{63.8} & 54.8 & 31.7 & 23.6\\
\texttt{HW} $\rightarrow$ \texttt{SIMP}  & 70.7 & 61.3 & 55.1 & \textbf{36.9} & 23.3\\
\midrule
\textbf{256M} & & & & &\\
\texttt{BASELINE}            & \textbf{73.8} & \textbf{65.6} & 55.0 & 30.1 & 23.5\\
\texttt{INTERLEAVED}            & 73.6 & 64.1 & \textbf{56.2} & \textbf{35.0} & 24.5\\
\texttt{SIMP} $\rightarrow$ \texttt{HW}  & 73.7 & 64.3 & 55.9 & 30.8 & 25.8\\
\texttt{HW} $\rightarrow$ \texttt{SIMP}  & 73.1 & 64.3    & 56.0 & 34.1 & \textbf{26.2}\\
\bottomrule
\end{tabularx}
\caption{
Zero-shot evaluations on linguistic competence (BLiMP, BLiMP-Supplement), world knowledge (EWoK), discourse (Entity Tracking), and general-domain reasoning (MMLU). 
For 124M models, \texttt{SIMP} $\rightarrow$ \texttt{HW} leads on BLiMP, while \texttt{HW} $\rightarrow$ \texttt{SIMP} achieves the highest score on Entity Tracking. 
At 256M, performance differences are narrower, with small trade-offs across tasks.
}
\label{tab:ling-benchmarks}
\end{table}

\subsection{Curriculum vs. Interleaving}
We next ask whether ordering simplified and human-written text into two distinct phases (curriculum or anti-curriculum) provides advantages over mixing them uniformly. Here, \texttt{INTERLEAVED} serves as the natural baseline, representing random shuffling of the two corpora.

\paragraph{Fine-tuning evaluation.}
At 124M, the curriculum schedule \texttt{SIMP}$\rightarrow$\texttt{HW} edges out \texttt{INTERLEAVED} across all budgets (+0.3, +0.1, +0.4), while \texttt{HW}$\rightarrow$\texttt{SIMP} is equal or slightly worse (Table~\ref{tab:nlu-summary}). The margins are small, but the consistent advantage for \texttt{SIMP}$\rightarrow$\texttt{HW} suggests that a warm-up on simplified text may provide downstream benefits when model capacity is limited. At 256M, the picture changes: \texttt{INTERLEAVED} remains strongest across all budgets, while the ordered schedules are competitive but do not surpass it (\texttt{SIMP}$\rightarrow$\texttt{HW}: $-0.4$, $-1.2$, $-1.3$ vs.\ \texttt{INTERLEAVED}; \texttt{HW}$\rightarrow$\texttt{SIMP}: $-0.2$, $-1.6$, $-2.1$). In short, simple-to-complex ordering helps at 124M, especially in low-resource settings where it shows improved sample efficiency, but offers no advantage as model capacity increases.

\paragraph{Zero-shot evaluation.}
For zero-shot tasks, ordering effects are somewhat stronger than in the fine-tuning results. At 124M, \texttt{SIMP}$\rightarrow$\texttt{HW} leads on BLiMP and BLiMP-Supplement, while \texttt{HW}$\rightarrow$\texttt{SIMP} excels on entity tracking (Table~\ref{tab:ling-benchmarks}). 
At 256M, however, \texttt{INTERLEAVED} typically matches or exceeds the ordered setups—for example, it leads on entity tracking (+4.3 vs.\ \texttt{SIMP}$\rightarrow$\texttt{HW}) and holds a small edge on HellaSwag (Table~\ref{tab:reasoning-benchmarks}). These patterns suggest that ordered exposure can steer smaller models toward specific strengths, but that random mixing is safer once capacity is sufficient.

\paragraph{Summary.}
Compared to interleaving, \texttt{SIMP} $\rightarrow$ \texttt{HW} yields modest gains at 124M—especially for NLU fine-tuning and linguistic probes—but these benefits diminish or reverse at 256M. This pattern suggests that smaller models benefit from a simple-to-complex progression in surface-level complexity, whereas larger models can internalize both variants without explicit ordering. \textbf{We view these as hypotheses under our training regime: ordering appears to matter more in low-capacity settings and is less critical when models scale up.}

\section{Conclusion}
\label{sec:conclusion}
This work examines whether augmenting pretraining data with LLM-based simplifications and ordering by text complexity improves representation quality in data-constrained settings. Through controlled data and training conditions, we isolated the effects of text complexity in curriculum design across two model sizes.

\break


\paragraph{Our findings suggest three takeaways:}
\begin{enumerate}
  \item[(1)] Adding simplified data outperforms repeating human-written text, yielding modest gains at 124M and clearer benefits at 256M.
  \item[(2)] Curriculum effects depend on model scale: smaller models benefit slightly more from a simple-to-complex curriculum, while larger models favor balanced exposure via interleaving.
  \item[(3)] Differences are most evident in zero-shot and low-resource fine-tuning scenarios, where schedule choice impacts representational quality, improving zero-shot performance and sample efficiency in small fine-tuning budgets.
\end{enumerate}

Overall, our results show that in data-constrained settings, text simplification and curriculum learning can complement repeated exposure. Although the gains are modest, they suggest practical ways to extend the utility of pretraining without fresh data collection. Future work includes testing longer training horizons, exploring rewriting methods beyond simplification, and evaluating schedule effects at larger scales and in non-English domains.

\section*{Limitations}
Our study is intentionally narrow in scope. We test only two model sizes (124M and 256M) with a single decoder-only architecture, leaving open how schedule effects scale to larger models or alternative designs. Training is constrained to a fixed budget of roughly 4B tokens (two full passes over a 2B-token human-written corpus), so results may differ under longer horizons or larger-scale training. We also study only one form of rewriting (LLM-based simplification) while other transformation types (e.g., paraphrasing, elaboration, style transfer) may behave differently. Finally, the evaluation focuses on educational data in English and a limited set of NLU and zero-shot benchmarks, the results may not transfer directly to other domains, languages, or broader task families.

\bibliography{custom}

\pagebreak
\appendix

\section{Per-task Results of Fine-tuning Evaluation}
Results are presented by fine-tuning budget: Full (Table \ref{tab:per-task-full}), Small (Table \ref{tab:per-task-small}), and Tiny (Table \ref{tab:per-task-tiny}).

\label{sec:per-task-tables}

\begin{table*}[t]
\centering
\footnotesize
\begin{tabularx}{\textwidth}{@{}Xcccccc@{}}
\toprule
\textbf{Model} & 
\multicolumn{1}{r}{\textbf{BoolQ}} & 
\multicolumn{1}{r}{\textbf{MNLI}} & 
\multicolumn{1}{r}{\textbf{MRPC}} & 
\multicolumn{1}{r}{\textbf{QQP}} & 
\multicolumn{1}{r}{\textbf{RTE}} & 
\multicolumn{1}{r}{\textbf{Avg.}} \\
\midrule

\textbf{124M} & & & & & & \\
\texttt{BASELINE}            & 70.6 ± 0.3 & 66.2 ± 0.7 & 82.2 ± 1.7 & 80.9 ± 0.3 & 66.4 ± 3.1 & 70.9 \\
\texttt{INTERLEAVED}         & 69.5 ± 0.6 & 67.5 ± 0.2 & 80.6 ± 1.0 & 81.1 ± 0.3 & 69.2 ± 2.4 & 71.1 \\
\texttt{SIMP} $\rightarrow$ \texttt{HW}  & 70.7 ± 0.6 & 67.1 ± 0.7 & 82.2 ± 2.2 & 81.3 ± 0.1 & 68.1 ± 2.8 & 71.4 \\
\texttt{HW} $\rightarrow$ \texttt{SIMP}  & 71.2 ± 0.7 & 66.7 ± 0.7 & 83.3 ± 0.7 & 80.6 ± 0.1 & 66.2 ± 1.7 & 71.1 \\
\midrule
\textbf{256M} & & & & & & \\
\texttt{BASELINE}            & 71.2 ± 0.1 & 69.2 ± 0.5 & 81.9 ± 1.1 & 81.5 ± 0.1 & 65.7 ± 3.1 & 73.9 \\
\texttt{INTERLEAVED}         & 71.0 ± 0.3 & 69.8 ± 0.1 & 84.1 ± 1.7 & 82.2 ± 0.1 & 71.5 ± 1.8 & 75.7 \\
\texttt{SIMP} $\rightarrow$ \texttt{HW}  & 71.0 ± 0.8 & 70.3 ± 0.3 & 82.9 ± 1.0 & 82.0 ± 0.1 & 70.1 ± 2.5 & 75.3 \\
\texttt{HW} $\rightarrow$ \texttt{SIMP}  & 71.8 ± 0.2 & 69.5 ± 0.3 & 82.2 ± 1.0 & 82.0 ± 0.2 & 71.8 ± 2.6 & 75.5 \\
\bottomrule
\end{tabularx}
\caption{Per-task fine-tuning results under \textbf{Full} fine-tuning budget. “Avg.” is the mean across tasks. Results are grouped by model size (124M and 256M) and curriculum strategy.}
\label{tab:per-task-full}
\end{table*}

\begin{table*}[t]
\centering
\footnotesize
\begin{tabularx}{\textwidth}{@{}Xcccccc@{}}
\toprule
\textbf{Model} & 
\multicolumn{1}{r}{\textbf{BoolQ}} & 
\multicolumn{1}{r}{\textbf{MNLI}} & 
\multicolumn{1}{r}{\textbf{MRPC}} & 
\multicolumn{1}{r}{\textbf{QQP}} & 
\multicolumn{1}{r}{\textbf{RTE}} & 
\multicolumn{1}{r}{\textbf{Avg.}} \\
\midrule

\textbf{124M} & & & & & & \\
\texttt{BASELINE}            & 62.9 ± 1.0 & 63.1 ± 0.4 & 70.8 ± 1.0 & 78.1 ± 0.9 & 63.4 ± 4.5 & 67.7 \\
\texttt{INTERLEAVED}         & 62.4 ± 0.3 & 64.0 ± 0.3 & 72.4 ± 1.0 & 78.7 ± 0.0 & 63.7 ± 1.6 & 68.2 \\
\texttt{SIMP} $\rightarrow$ \texttt{HW}  & 64.1 ± 0.6 & 63.1 ± 0.3 & 71.5 ± 0.7 & 79.0 ± 0.2 & 63.9 ± 0.7 & 68.3 \\
\texttt{HW} $\rightarrow$ \texttt{SIMP}  & 62.7 ± 0.7 & 63.1 ± 0.2 & 73.2 ± 0.3 & 78.3 ± 0.4 & 60.2 ± 3.3 & 67.5 \\
\midrule
\textbf{256M} & & & & & & \\
\texttt{BASELINE}            & 63.9 ± 0.7 & 65.1 ± 0.2 & 72.8 ± 2.3 & 79.3 ± 0.1 & 63.2 ± 2.4 & 68.8 \\
\texttt{INTERLEAVED}         & 63.7 ± 0.6 & 65.8 ± 0.5 & 73.1 ± 1.7 & 80.1 ± 0.2 & 71.1 ± 2.8 & 70.7 \\
\texttt{SIMP} $\rightarrow$ \texttt{HW}  & 64.2 ± 0.3 & 66.1 ± 0.7 & 71.5 ± 1.7 & 79.7 ± 0.1 & 66.0 ± 0.7 & 69.5 \\
\texttt{HW} $\rightarrow$ \texttt{SIMP}  & 63.7 ± 0.2 & 65.4 ± 0.7 & 71.6 ± 3.8 & 79.4 ± 0.2 & 65.3 ± 4.9 & 69.1 \\
\bottomrule
\end{tabularx}
\caption{Per-task fine-tuning results under \textbf{Small} fine-tuning budget. “Avg.” is the mean across tasks. Results are grouped by model size (124M and 256M) and curriculum strategy.}
\label{tab:per-task-small}
\end{table*}

\begin{table*}[t]
\centering
\footnotesize
\begin{tabularx}{\textwidth}{@{}Xcccccc@{}}
\toprule
\textbf{Model} & 
\multicolumn{1}{r}{\textbf{BoolQ}} & 
\multicolumn{1}{r}{\textbf{MNLI}} & 
\multicolumn{1}{r}{\textbf{MRPC}} & 
\multicolumn{1}{r}{\textbf{QQP}} & 
\multicolumn{1}{r}{\textbf{RTE}} & 
\multicolumn{1}{r}{\textbf{Avg.}} \\
\midrule

\textbf{124M} & & & & & & \\
\texttt{BASELINE}            & 59.5 ± 0.7 & 59.4 ± 1.0 & 67.9 ± 0.3 & 75.3 ± 0.4 & 59.7 ± 6.1 & 64.4 \\
\texttt{INTERLEAVED}         & 58.4 ± 0.4 & 60.7 ± 0.3 & 70.4 ± 1.5 & 76.0 ± 0.2 & 58.3 ± 1.2 & 64.8 \\
\texttt{SIMP} $\rightarrow$ \texttt{HW}  & 58.8 ± 1.0 & 59.8 ± 0.4 & 68.9 ± 2.6 & 76.5 ± 0.1 & 62.0 ± 0.8 & 65.2 \\
\texttt{HW} $\rightarrow$ \texttt{SIMP}  & 59.5 ± 1.5 & 60.3 ± 0.1 & 71.6 ± 0.8 & 75.6 ± 0.5 & 55.1 ± 3.3 & 64.4 \\
\midrule
\textbf{256M} & & & & & & \\
\texttt{BASELINE}            & 60.8 ± 1.3 & 60.7 ± 0.4 & 68.1 ± 1.2 & 76.5 ± 0.6 & 59.0 ± 1.8 & 65.0 \\
\texttt{INTERLEAVED}         & 60.1 ± 0.7 & 61.6 ± 0.5 & 71.0 ± 1.9 & 77.5 ± 0.2 & 64.6 ± 0.7 & 66.9 \\
\texttt{SIMP} $\rightarrow$ \texttt{HW}  & 60.5 ± 1.0 & 60.9 ± 0.9 & 68.6 ± 1.8 & 76.9 ± 0.3 & 61.1 ± 2.5 & 65.6 \\
\texttt{HW} $\rightarrow$ \texttt{SIMP}  & 59.9 ± 0.6 & 61.4 ± 0.4 & 65.5 ± 4.1 & 76.2 ± 0.3 & 60.9 ± 4.2 & 64.8 \\
\bottomrule
\end{tabularx}
\caption{Per-task fine-tuning results under \textbf{Tiny} fine-tuning budget. “Avg.” is the mean across tasks. Results are grouped by model size (124M and 256M) and curriculum strategy.}
\label{tab:per-task-tiny}
\end{table*}

\end{document}